# SALIENCY DETECTION FOR IMPROVING OBJECT PROPOSALS

*Shuhan Chen*[†‡], *Jindong Li*[†], *Xuelong Hu*[†], *Ping Zhou*[‡]

[†]College of Information Engineering, Yangzhou University, Yangzhou, China
[‡]Wanfang Electronic Technology Co., Ltd., Yangzhou, China

**ABSTRACT**

Object proposals greatly benefit object detection task in recent state-of-the-art works, such as R-CNN [2]. However, the existing object proposals usually have low localization accuracy at high intersection over union threshold. To address it, we apply saliency detection to each bounding box to improve their quality in this paper. We first present a geodesic saliency detection method in contour, which is designed to find closed contours. Then, we apply it to each candidate box with multi-sizes, and refined boxes can be easily produced in the obtained saliency maps which are further used to calculate saliency scores for proposal ranking. Experiments on PASCAL VOC 2007 test dataset demonstrate the proposed refinement approach can greatly improve existing models.

## 1. INTRODUCTION

Benefit from the success of object proposal methods (e.g. [1]), which selects an ideal number of region proposals to cover most observable objects, object detection has made a great progress in recent years, such as region-based convolutional neural networks (R-CNN) [2], Fast R-CNN [3]. A good proposal generator should efficiently output candidate bounding boxes as few as possible to reach recall rate as high as possible.

As summarized in [4], the existing object proposal generators can be roughly classified into two classes: grouping based [1, 6-9] and sliding window based [10-12]. The first type approaches usually generate relatively high accurate object proposals but require long computation time [4]. While the later typically produce proposals efficiently due to be independent of superpixel segmentation but with low localization accuracy [4]. Most of the above methods achieve high recall at the cost of sampling a large number of candidate boxes, which will prevent computationally expensive classifiers to be applied in subsequent process (e.g. object detection). To address it, an accurate objectness score or some other improvement is needed. In [13], each candidate box was scored by various objectness cues including color, contrast, edge and saliency map. A linear classifier was used to calculate the objectness score in [10]. In [11], objectness score was computed by counting the number of edges that are wholly contained in a candidate box, while in [9] edge was used as its summation on outer boundary and then normalized by its length. Contour score [14] was proposed by combing completeness and tightness to reject non-object proposals. In [15], superpixel tightness was used as a localization bias indicator which also can be applied for proposal ranking. Non-maximal suppression (NMS) [16] is also an effective way to reduce proposal redundancy [11, 15]. Although such progress have been made in this young field, it is still very challenging to generate object proposals with high quality which means high recall with few proposals at high intersection over union (IoU) threshold.

Saliency detection is a very active research area recently and has been successfully applied in various applications [26-27], which aims to make certain objects or regions of an image stand out from their neighbors [17]. We find that object proposal generation and saliency detection are high related and can be fused to improve each other as described in [18]. Different with [18], which presented a graphical model by iteratively optimizing a novel energy function to integrate these two aspects, we directly apply saliency detection to each candidate box to refine them based on the following observation: a good candidate box should contain only one wholly object centered in it, which can be seen as an inside salient object that can be well captured by the existing bottom-up saliency detection methods. Although we can efficiently detect salient objects in one image, it will meet computation bottleneck when dealing with thousands of bounding boxes per image. To address it, we apply geodesic saliency [19] in contour image to detect salient object for each candidate box. It can be more efficient since the feature construction in contour is very simple compared with color images for saliency detection. More than that, obtained saliency map is also used to calculate a saliency score for proposal re-ranking. To introduce high diversity, multi-size windows are further selected for saliency detection. We test it on the Pascal VOC 2007 test dataset [25] and achieve great improvement in the existing methods especially at high IoU. In particular, we achieve the highest recall at IoU of 0.8 with 64.8% by using 1000 proposals.

## 2. THE PROPOSED METHOD

In this section, we describe our method to improve object proposals based on saliency detection in contour. The initial boxes are provided by the existing approaches, then we apply geodesic saliency metric in contour to detect saliency regions in each candidate box with multi-sizes. Based on the obtained saliency maps, refined boxes can be obtained in the segmentation results and then ranked by their saliency scores. The pipeline of the proposed refinement is shown in Fig.1.

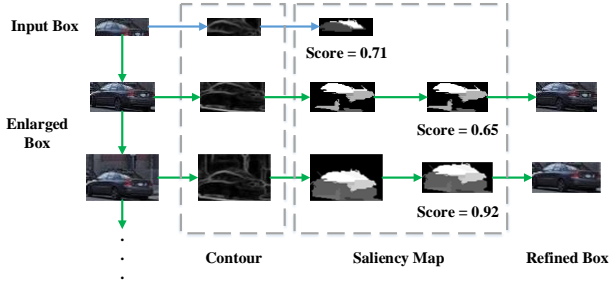

**Fig.1.** The pipeline of the proposed method.

### 2.1. Geodesic Saliency in Contour

Saliency detection aims to detect the most salient objects (usually only one) in an image. While object proposal detector attempts to output a reasonable number of bounding boxes to cover most objects in an image, and each of them only covers one which can be seen as a salient object in it intuitively. Many effective bottom-up saliency detection approaches have been proposed in the literatures [21]. One of the most effective cues is backgroundness which assumes four borders of an image as pseudo-background, and then saliency can be measured by the contrast versus this background [19]. However, it may be noisy when object touches one or more image borders. In such case, it is still very difficult to get pure background border regions. Fortunately, we needn't worry about that in box saliency detection. In object proposal generation, a candidate box is inaccurate to localize an object if it straddles the box borders. Thus, we can simply select the regions surrounding the box as its background priors. Then, saliency can be simply measured by computing the difference between them as common saliency detection does. The candidate box straddled by its inside object will have low saliency value, which indicates low objectness. In contrast, high saliency value denotes high objectness. Thus, obtained saliency map can be intuitively utilized to score its corresponding box.

Salient objects usually have distinctive colors, to capture it, CIE LAB is the most used color space. To get high visual quality, diffusion or propagation based methods are further explored recently [22-23]. However, it will meet computation bottleneck if we directly apply these complex approaches into object proposals due to the huge number. To this end, we make saliency detection in contour which can be generated by the efficient algorithm in [20]. Note that it has not been explored in saliency detection as far as we know. Contour is an effective cue for proposal generation and has been successfully applied in some previous works [11, 14], which are based on the observation that the box with a closed contour is more likely to be an object. Different with them, we measure it by saliency detection.

In box saliency detection, backgroundness can be re-explained as the background regions are connected to the box borders while it is hard for the objects. This prosperity can be well captured by the effective and efficient geodesic distance metric which has been successfully applied in color image saliency detection [19], where regional saliency is measured by the length of its shortest path to the virtual background node in a weighted graph.

Based on the above observations, saliency detection in contour can well capture the probability of a bounding box containing an object. Specifically, given an image, graph-based segmentation [24] is first used to segment it into superpixels, and then structured forests algorithm [20] to produce its contour. The distance between a joint superpixel pair is measured by summing up the contour responses within the common boundary pixels and normalized by the length of the common boundary as described in [9]. Let $C(z)$ to be the contour response at pixel $z$, and $l(i, j)$ denote the common boundary pixels set, then the distance between superpixel $i$ and $j$ is defined as:

$$D(i,j) = \begin{cases} \frac{1}{|l(i,j)|} \sum_{z \in l(i,j)} C(z) & |l(i,j)| \neq 0 \\ 0 & \text{otherwise} \end{cases} \quad (1)$$

Based on the above distance metric, the geodesic distance can be formulated as the accumulated distances along the shortest path from $t$ to background superpixels $B$. Let $\pi = \{\pi(0),\ldots, \pi(K)\}$ be the shortest path in the set $\Pi_{B,t}$ of all paths, saliency for superpixel $t$ can be measured as:

$$S(t) = \min_{\pi \in \Pi_{B,t}} \sum_{p=1}^{K} D(\pi(p), \pi(p-1)) \quad (2)$$

### 2.2. Box Refinement

To improve the localization accuracy of the candidate boxes, we present a box refinement method using the above geodesic saliency detection. Specifically, given a bounding box $b$, we first need to select its background superpixels. Here, we simply choose its four borders with single pixel width. Based on it, we can easily get a saliency map for the given box using our geodesic saliency detection. Note that each saliency map is normalized into [0, 1] for refinement. Some detection results are shown in Fig.2, it can be clearly observed that object inside each box can be well highlighted by our proposed approach. Then, a refined box $b'$ can be obtained from the binarized saliency map with a threshold $T$.

Superpixels in a box with low localization accuracy usually straddle box borders, which indicates low tightness as mentioned in [13, 15]. To refine these boxes, we need to enlarge them to include more image contents for saliency detection to capture the whole object. Thus, we introduce multi-size windows to solve it. Specifically, totally $M$ enlarged windows with $\delta$ pixels width are selected for each candidate box. By this means, $M$ new boxes are generated for each initial candidate. As shown in Fig.1, considering multi-size windows can well refine these inaccurate boxes. However, more redundancy will also be introduced thus need to be reduced to a moderate budget of proposals.

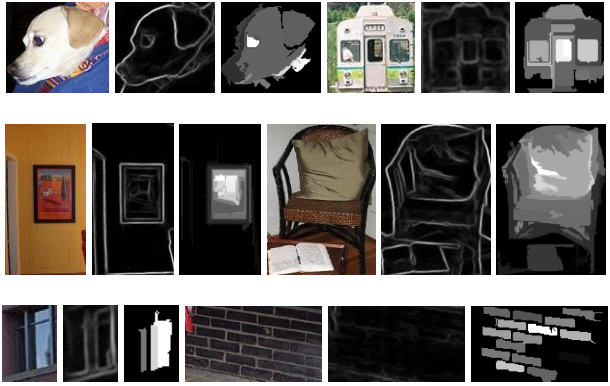

**Fig.2.** The proposed geodesic saliency detection results of different object proposals.

### 2.3. Proposals Re-Ranking

As mentioned above, the refined boxes may contain a large number of redundancy. Thus, we define a ranking function to prune them. Generally, an accurate object box typically has high saliency values, while it is not for a background box. Furthermore, in our multi-sizes saliency detection, the larger the window is, the less confident the refined proposal will be, that's because more included contours will introduce more noisy saliency detection results. Considering them together, we define our ranking function for each box $b$ as:

$$score_b^m = \frac{(M+1-m)\sum_{t\in b}S(t)\cdot size(t)}{|b|^\lambda} \quad (3)$$

where $|b|$ is the area of $b$, $m = 0,1,\ldots,M$, and $\lambda$ is set to be less than 1 to favor larger boxes. Then, all the proposals are sorted in descending order.

Nevertheless, saliency detection results may be noisy when the extracted contours are not accurate enough, which will lead to inaccurate ranking orders. To improve it, we combine the refined box sets together with the input sets for ranking to get our final proposals. In detail, each of them are re-scored by its normalized inverse indexes and then re-sorted in descending order in the combined sets. Finally, NMS is performed to obtain the final proposals as did in [11, 15] by setting the IoU to 0.9.

## 3. EXPERIMENTS

We evaluate our method on the PASCAL VOC2007 dataset [25] which contains 9,963 images spread over 20 categories, and each of them has a bounding box for each object. We only test it on the test set as previous works. The common recall metric is used to evaluate the quality of object proposals in this paper following [4], including recall-IoU and recall-proposal curves. In addition, average recall (AR) between IoU 0.5 to 1.0 is further computed to measure the overall accuracy of proposals. In our experiments, $T$ and $\lambda$ are set to 0.01 and 0.9 respectively, $\sigma = 0.8$, and $k = 100$ for graph-based segmentation, all the parameters are unchanged in the following experiments.

### 3.1. Influence of Window Sizes

We first examine the performance of the proposed method by using different sizes in saliency detection. In this experiment, we compare the results by varying $M$ from 0 (single size) to 4 using MCG (Multiscale Combinatorial Grouping) [7] as the initial candidates, in which $\delta = \{1\}$, $\{1, 5\}$, $\{1, 5, 15\}$, $\{1, 5, 15, 25\}$, and $\{1, 5, 15, 25, 40\}$, respectively. The number of proposals tested is 2000 and the results are shown in Fig.3. As can be seen, best recall rate is obtained when $M = 3$ nor 4. That's not surprise because larger window will introduce more noisy results as mentioned before. Thus, $M$ is fixed to 3 in the following experiments. We also find that our method is not very sensitive to $M$. In generally, more sizes are needed for the methods with poor localization accuracy in our refinement.

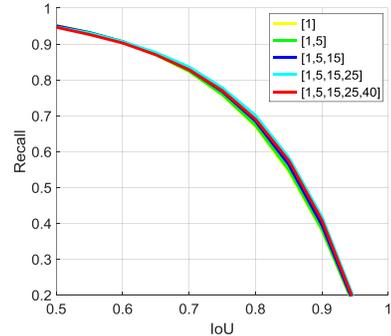

**Fig.3.** Evaluation of performance with different sizes using 2000 proposals.

### 3.2. Validation of the Proposed Approach

In this section, we verify the effectiveness of the proposed refinement with the existing models, including SS (Selective Search) [1], EB (Edge Boxes 70) [11], and MCG [7]. We also report the performance of a recent published improving method MTSE [15] for comparison. The variants of our saliency refinement integrated models are recommended as S-SS, S-EB, S-MCG, and M-SS, M-EB, M-MCG for MTSE. Fig.4 (a)-(c) show their performances using recall-IoU curve (500, 1000, and 2000 proposals), we can clearly see that our saliency refinement successfully improves the existing methods by a large margin and even performs slightly better than MTSE when IoU higher than 0.7. The improvement at low IoU is not so significant, e.g. EB is only boosted at high

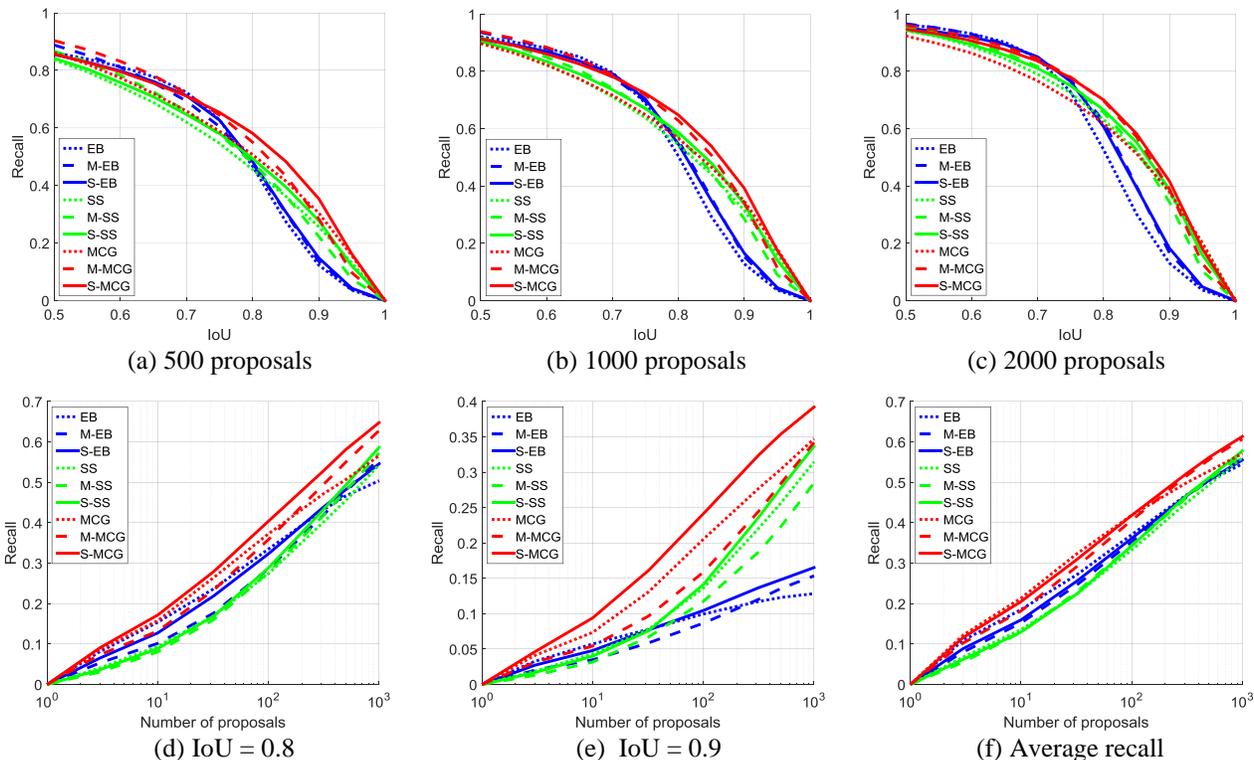

**Fig.4.** Performance of different methods and their improved versions by MTSE and our saliency refinement.

IoU by our refinement. It can be explained that the localization of the box with low IoU is too coarse for saliency detection to refine it even with multi-sizes. It also can be verified by the observation that the improvements in EB and SS are not comparable with MCG especially in the case of small number proposals. In particular, S-MCG achieves best 64.8% recall at strict 0.8 IoU when using 1000 proposals. We also consider the challenge IoU 0.8 even 0.9 and AR by varying the number of proposals from 1 to 1000 shown in Fig.4 (d)-(f). Again, performance boost can be observed by our saliency refinement, in which S-MCG still achieves highest recall in all the cases. Therefore, we believe that our saliency refinement can benefit object detection task with better localization.

### 3.3. Failure Cases

Our saliency refinement is based on the saliency detection results which is determined by the accuracy of contour detection. Thus, it will get worse refined proposals if the contour of the object is weak or even missing. Another we need to point out is that our approach is trying to find the closed contour to indicate objectness, which is similar with [14]. Therefore, it tends to pop out some box with closed contour while is not a semantic object, such as windows or some small structures. In other words, there are still some non-object proposals with high objectness scores need to be removed, which is our future work. Some failure cases are shown in the last row of Fig.2.

### 4. CONCLUSIONS

In this paper, we present a new geodesic saliency detection in contour to improve the quality of object proposals. Based on the obtained saliency map of each candidate box, we get a refined box with better localization than the initial one. By further applying multi-sizes in saliency detection, the input candidate is well refined with both high diversity and accurate localization. Finally, boxes with high objectness are pop out by their saliency scores. By integrating our saliency refinement, all the existing methods are improved by a large margin in PASCAL VOC 2007 test dataset both at high IoU threshold and few proposal number, which can benefit subsequent object detection task. It is also worthy to note that our geodesic saliency detection in contour can be directly applied for salient object detection task. We also wish to see more related works to be explored in this direction. In the future, we will first try to accelerate our code by re-writing in C/C++. Another concern is to apply more powerful contour detection method to improve performance. The source code will be public available for research purpose after publication by email request.

**Acknowledgements** This work was supported by Foundation of Jiangsu Educational Committee (15KJB510032).